\documentclass[graybox]{template/svmult}

\usepackage{mathptmx}       
\usepackage{helvet}         
\usepackage{courier}        
\usepackage{type1cm}        
\usepackage{wrapfig, url}
\usepackage{siunitx}
\usepackage{algpseudocode}
\usepackage{algorithm}
\usepackage{makeidx}         
\usepackage{graphicx}        
\usepackage{multirow}
\usepackage{multicol}        
\usepackage[bottom]{footmisc}
\usepackage{hyphenat}

\usepackage[numbers]{natbib}

\long\def\invis#1{}

\newcommand\etal{\textit{et al.\ }}

\makeindex             

\begin{document}

\title*{Dynamic Autonomous Surface Vehicle Controls Under Changing Environmental Forces}
\titlerunning{Dynamic ASV Controls} 
\author{Jason Moulton, Nare Karapetyan, Michail Kalaitzakis, Alberto Quattrini Li,  Nikolaos Vitzilaios, and Ioannis Rekleitis}
\authorrunning{Moulton, Karapetyan, Kalaitzakis, Quattrini Li, Vitzilaios, Rekleitis}
\institute{Jason Moulton, Nare Karapetyan, Ioannis Rekleitis \at Computer Science \& Engineering Department, University of South Carolina, \email{moulton, nare@email.sc.edu, yiannisr@cse.sc.edu}
\and Michail Kalaitzakis, Nikolaos Vitzilaios \at Department of Mechanical Engineering, University of South Carolina,
\email{michailk@email.sc.edu, vitzilaios@sc.edu}
\and Alberto Quattrini Li \at Computer Science Department, Dartmouth College, \email{alberto.quattrini.li@dartmouth.edu}}
%
%
\maketitle


\abstract{The ability to navigate, search, and monitor dynamic marine environments such as ports, deltas, tributaries, and rivers presents several challenges to both human operated and autonomously operated surface vehicles. Human data collection and monitoring is overly taxing and inconsistent when faced with large coverage areas, disturbed environments, and potentially uninhabitable situations. In contrast, the same missions become achievable with Autonomous Surface Vehicles (ASVs) configured and capable of accurately maneuvering in such environments. The two dynamic factors that present formidable challenges to completing precise maneuvers in coastal and moving waters are currents and winds. In this work, we present novel and inexpensive methods for sensing these external forces, together with methods for accurately controlling an ASV in the presence of such external forces. The resulting platform is capable of deploying bathymetric and water quality monitoring sensors. Experimental results in local lakes and rivers demonstrate the feasibility of the proposed approach.
\invis{In addition, this ASV is capable of hosting added payload for protective equipment when deploying in areas deemed too dangerous for human exposure.}}

\section{Introduction}
\label{sec:intro}

\begin{wrapfigure}[13]{r}{0.5\textwidth}
\centering
\vspace{-0.3in}
\includegraphics[width=0.5\textwidth]{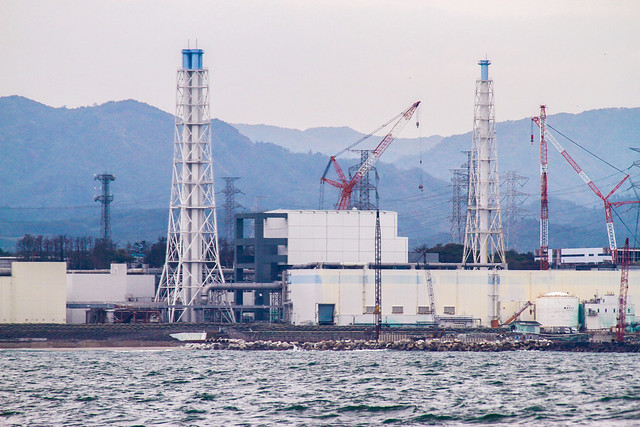}
\caption{\label{fig:fuku}Fukushima Daiichi nuclear complex in Okumamachi, Japan taken during water sampling and analysis by IEAE in 2013.}
\end{wrapfigure}

As the demand for data collection and monitoring continues to expand across all reaches of the globe, research and development of Autonomous Surface Vehicles (ASVs) control in uncertain environments is essential. While the tasks and missions to which an ASV could be assigned are only limited by one's imagination, our desire to explore the unexplored increases the capabilities required in an ASV. One such hypothetical employment for  an ASV would have been to assist with monitoring and recovery after the Fukushima Daiichi nuclear disaster following the 2011 earthquake in Japan (Figure \ref{fig:fuku})\footnote{\url{https://www.flickr.com/photos/iaea_imagebank/10722882954}}.

For a less catastrophic scenario, with over 3.5 million miles of rivers in the United States alone, the ability to access, cover, and navigate them requires an ASV with long range potential, as well as a precise trajectory following capability to ensure safe maneuvers. 
In addition, the ability to take into account the effect of external forces would improve the efficiency in planning for coverage as well as savings in power and fuel consumption. While there exists much research into the effects of natural phenomena such as wind and current in ocean areas, there remains a void when it comes to studying the same type of effects on smaller ASVs in confined areas with higher currents such as rivers. Operating in the air and water domains simultaneously exposes ASVs to wind and current external forces that can easily overwhelm current Proportional, Integral, Derivative (PID) controlled navigation systems. 

Our paper pushes the research boundaries to advance the state of the art which will allow ASVs to be utilized in increasingly challenging conditions to ensure that ASVs become ubiquitous with researchers, engineers, and environmental scientists. 

\subsection{Problem Definition}
\label{subsec:problem}

\begin{wrapfigure}[12]{r}{0.5\textwidth}
\vspace{-2em}
\centering
\includegraphics[width=0.5\textwidth]{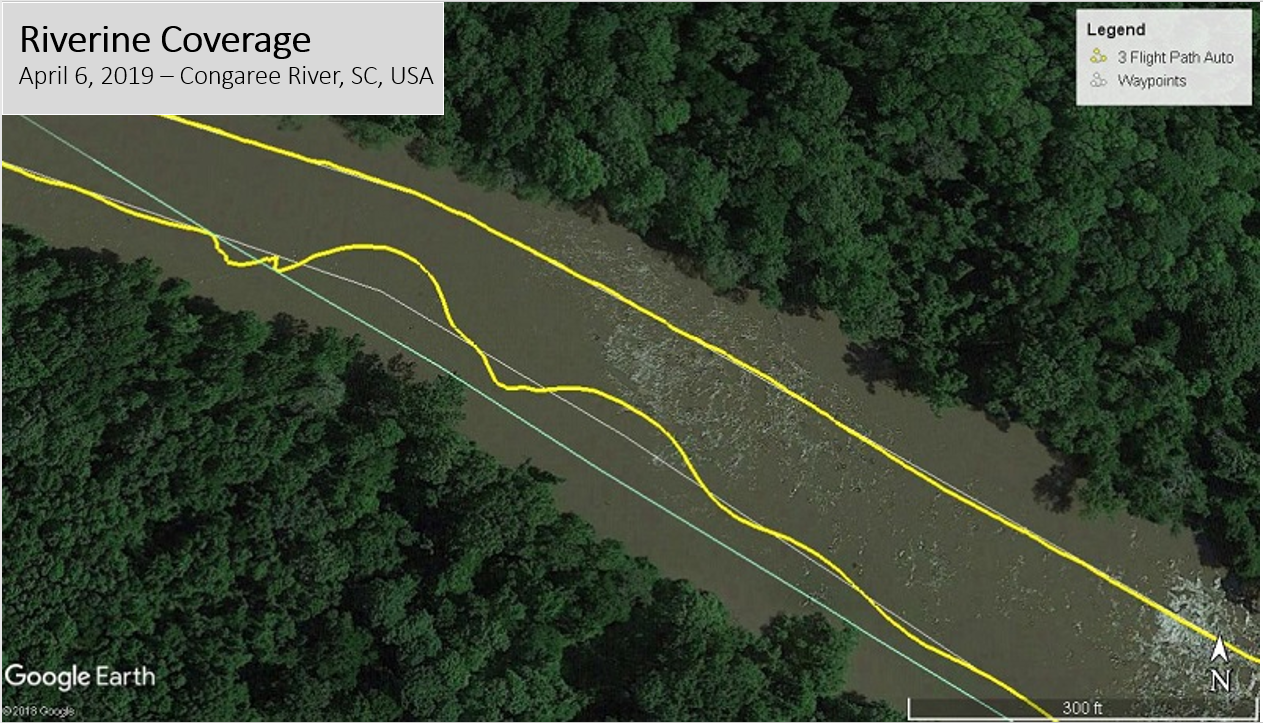}\vspace{-0.05in}
\caption{\label{fig:problem}Target trajectory unable to be followed in the downstream pass due to high \SI{3.0}{\m/\s} currents in the area.}
\end{wrapfigure}

Addressing the challenge of operating in the presence of non\hyp trivial external forces can be done in two different scales. If a long\hyp range map of the external forces is available then large scale planning can take the effects of the external forces into account. For example,  coverage planning algorithms~\cite{icra2018,karapetyan2019riverine}, can include the force map as an input variable in order to improve mission planning. In a smaller scale, real\hyp time force measurements can be used in a reactive controller to accurately track the desired trajectory. In analogy, knowing the traffic patterns in a city can generate routes through less congested streets, while a driver sensing slipping on ice, or pushed by a wind gust can guide the vehicle accordingly. In this paper we provide a novel method for augmenting a controller with information from local disturbances. A manifestation of this problem is illustrated in Figure \ref{fig:problem}, where the ASV is unable to maintain an accurate trajectory due to the PID controller being overcome by the changing currents. 

\invis{
The challenge of operating in dynamically changing environments can be approached on two fronts. If able to produce a long\hyp range map of the external forces applied to the vehicle in a given environment, then trajectory generations, such as coverage planning algorithms~\cite{icra2018,karapetyan2019riverine}, can include this map as an input variable in order to overall improve mission planning. 
Another front of this challenge is to adapt real-time to ever changing dynamics so that the ASV can maintain accurate path following to collect the required data at the desired locations. In this work, we present a novel approach to inexpensively solve the second issue,}

In most riverine environments, it should be noted that a standard PID driven way\hyp point navigation controller can be tuned to maintain course either when moving with or against the external force, but not both conditions with the same gains. See for example Figure \ref{fig:problem}, where the trajectory is followed accurately upstream, meaning against the current, and the erratic trajectory is produced from a downstream path.

\subsection{Related Work}
\label{subsec:rw}
There have been several approaches in developing small autonomous surface vehicles. Different combinations of single hull, twin\hyp hull, electric, or gas combustion can be observed in several publications \cite{mahacek2008development, GirdharIROS2011, fraga2014squirtle, curcio2005scout}.
Rodriquez \etal present a comprehensive feasibility review of ASVs as of 2012 \cite{rodriquez2012study}.
Recently, Woods Hole Oceanographic Institute (WHOI) \cite{whoiMokai2014} created an ASV based off Mokai's gas powered kayak. The capability for longer duration and increased payload led us to utilize the same base platform for our ASV~\cite{moulton2018autonomous}.

Rasal of Santa Clara University presented early work in 2013 on the  building of their SWATH ASV and applied some strategies for overcoming environmental disturbances based on their effects measured by an integrated IMU \cite{sclara}. While their results were improved over the standard way\hyp point navigator, control based on reactive measurements does not accomplish our goal of path following in faster moving currents and winds. However, their off-board control system inspires our design and implementation for customized control sequences for future mission\hyp specific tasks in longer\hyp range, more volatile environments. 

Using the platform and the integrated sensor suite presented in prior work \cite{icra2018,moulton2018autonomous}, we developed methods for measuring wind and current acting on the ASV. This led to the ability to predict wind and current forces using a Gaussian Process for short temporal periods. In addition, this collection of  measurements can enable potentially both optimal mission planning as well as discrete feed\hyp forward control development \cite{hsieh2018small}.  

Controller development research mainly relies on models using inertial\hyp tracking and compensating for roll, pitch, and yaw rates to provide course corrections. Our approach is more proactive, in that we actively measure and model environmental variables and provide the course correction prior to being swept off course.

Related research focused solely on minimizing error tracking includes work from Tsu-Chin \cite{tsao1994optimal} and robust digital tracking control based on a disturbance observer from Lee \etal \cite{lee1996robust}. These works closely model our setup, except that, in our case, the wind and current sensors are taking on the role of the disturbance observers. Pereira \etal focused on position anchoring of small under\hyp actuated ASVs in windy conditions \cite{pereira2008experimental}. Their performance was good in conditions the authors admitted to be moderate. So, in order to enable accurate tracking in volatile conditions, we will extend our tracking control research to be proactive. The two essential steps in completing this work are first, modeling the effects of winds and currents on an ASV, followed by using this model to implement countermeasures in the form of a feed\hyp forward controller to overcome them and maintain accurate path following. This is accomplished through over 60 deployments into a variety of conditions to measure the wind and current phenomena and develop a model, based on our observations, of the effects of external forces on the ASV's behavior. Finally, the effect\hyp based model feeds the improved autopilot controller to refine the ASV's navigation to overcome the currents and winds.

\subsection{Contributions}
\label{subsec:contributions}
\begin{figure}[h]
		\centering      
		\includegraphics[width=0.75\textwidth]{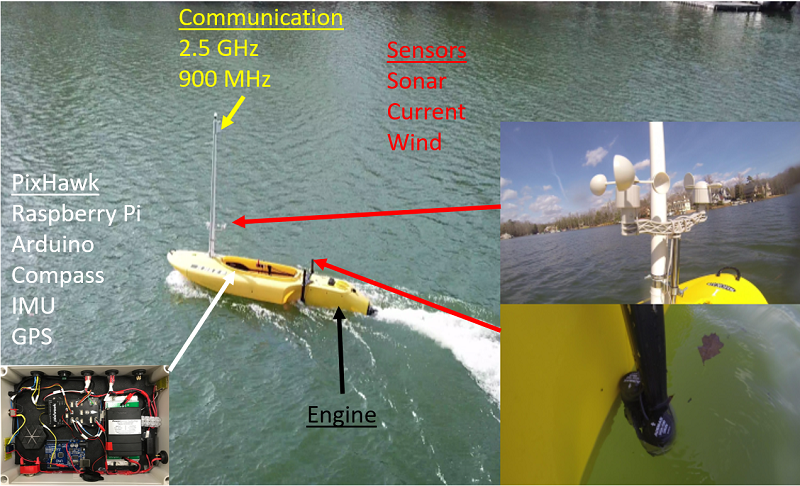}
		\caption{UofSC's environmental dynamic measurement platform with anemometer, current sensors, depth sounder, GPS, IMU, robust communications, and a ROS\hyp based data collection computer onboard.}
		\label{fig:jetyak2}
	\end{figure}

The contribution of this work is the augmentation of the ASV's current control system with feed\hyp forward controls to overcome the external dynamics and maintain a more accurate trajectory, by using measurements and models of natural disturbances affecting an ASV (Figure \ref{fig:jetyak2}) proposed in our previous work. Such a contribution will provide the greater scientific community with a more precise platform for data collection in challenging environments. In addition, it can provide an efficient and robust tool to aid search and rescue operators as well as environmental monitoring and bridge inspection teams.

The following section presents the methodology for accomplishing our goal as inexpensively as possible, with a brief discussion on the effects of external forces acting on an ASV, followed by a detailed proactive control augmentation description. Section \ref{sec:exp} presents our experimental setup and approach to create a field trial testing environment to produce meaningful results in Section \ref{sec:results}. Finally, we conclude with a short discussion of the results and suggestions for future work in this area.

\section{Methodology}
\label{sec:method}
This section describes the strategy we employ to solve the problem presented in Section \ref{subsec:problem}. For completeness, we will first briefly review the method for measuring external forces and modeling their effects on an ASV; for more information please see Moulton \etal \cite{moulton2018external}.
\invis{In order to do so, we must re-examine our implementation method for measuring external forces and modeling their effects on an ASV.} Then we present our approach of using these effects to implement proactive path-following control augmentation.

\begin{figure*}[ht]
		\centering
		\leavevmode
		\begin{tabular}{cc}
	        \includegraphics[width=0.5\textwidth]{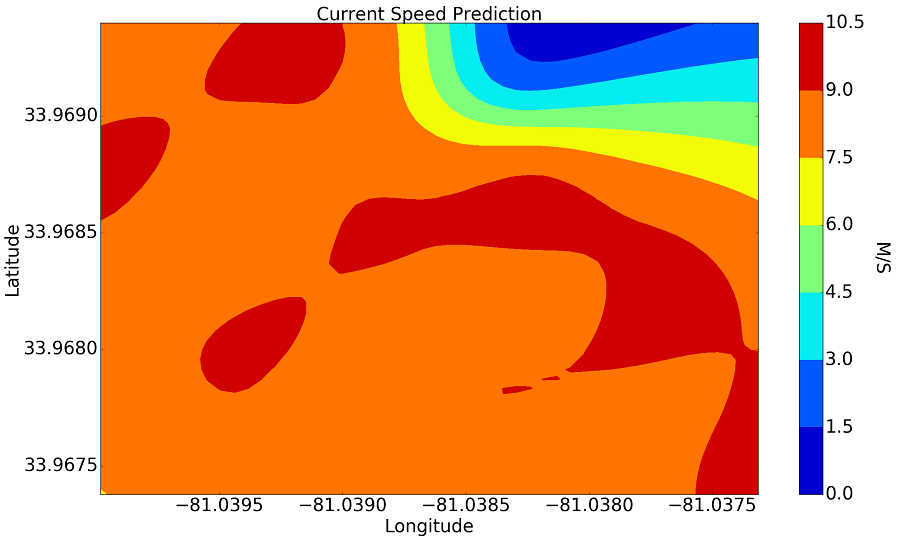}\label{fig:c1} &
			\includegraphics[width=0.5\textwidth]{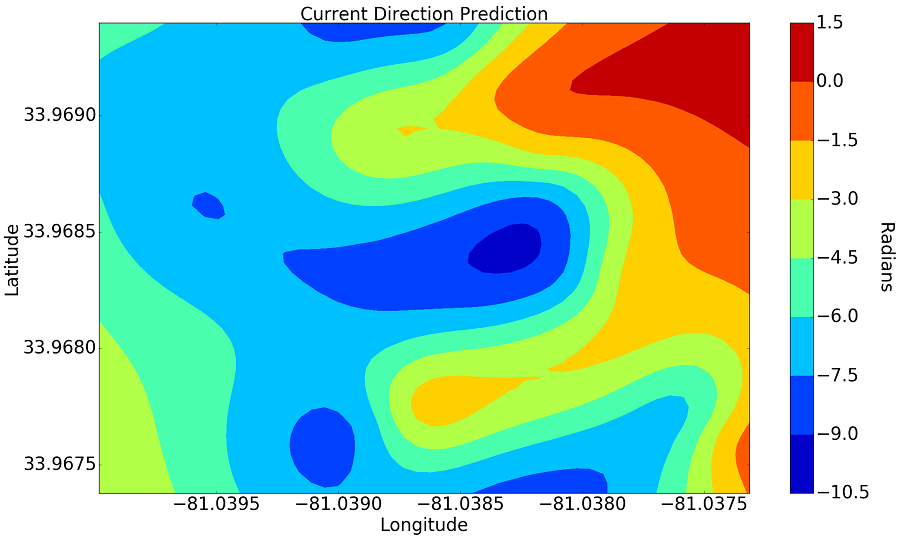}\label{fig:c2}
		\end{tabular}
		\caption{(a) Current speed prediction, (b) Current direction prediction during flood stage on Congaree River, SC. 
		\label{fig:current_predictions}}
\end{figure*}

\subsection{External Force Effects}
\label{subsec:effects}
\begin{figure*}[ht]
    \centering
    \leavevmode
    \begin{tabular}{cc}
        \includegraphics[width=0.49\textwidth]{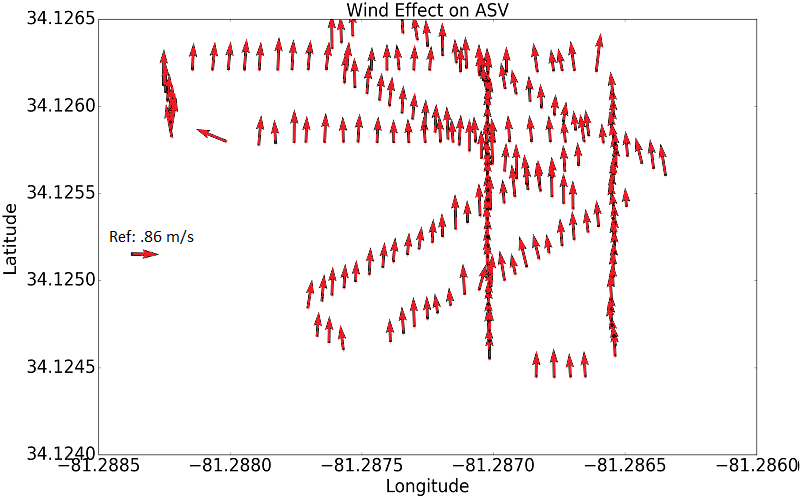}\label{fig:lake_effects} &
        \includegraphics[width=0.49\textwidth]{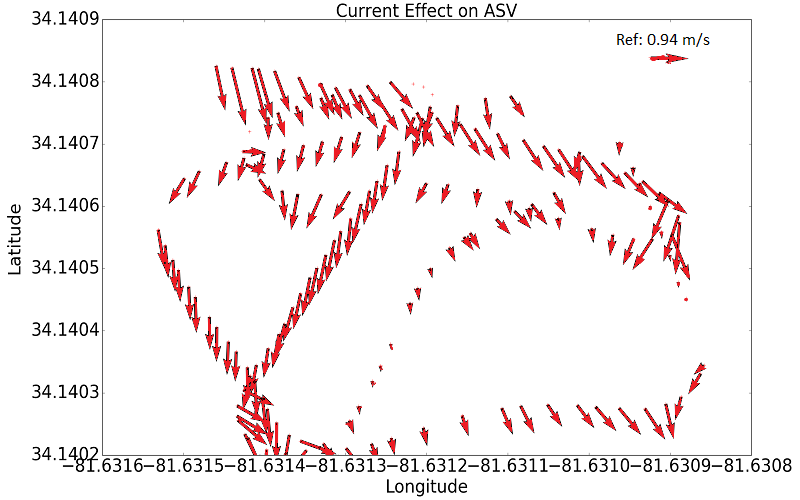}\label{fig:river_effects}
    \end{tabular}
    \caption{The effects of the wind and current on the ASV. Illustration reflects different scales due to the dominant effect of current over wind on the ASV.\label{fig:effects_map}}
\end{figure*}

There are two overlapping areas that benefit from measuring the external forces acting on the ASV. The first area, addressed in our prior work, is the ability to create a high\hyp level force map of a given phenomenon (see Figure \ref{fig:current_predictions}). This capability enables planning algorithms such as the one proposed by Karapetyan \etal \cite{icra2018} to pre\hyp select deployment sites and plan more efficient coverage solutions prior to launch. The second benefit results from the ability to use machine learning techniques for regression to produce effects models for the impact external forces are having on the robot. Figure \ref{fig:effects_map} illustrates the impact vector that the external force is having on the ASV. This capability enables the work presented in this paper, which in high\hyp level terms, the modeling of the effects feeds an adaptive controller which counteracts the external forces allowing for more accurate trajectory following of the ASV. 

\subsection{Proactive Control Through Way\hyp point Augmentation}
\label{subsec:control}
\begin{figure}[t]
\centering      
    \includegraphics[width=0.9\textwidth]{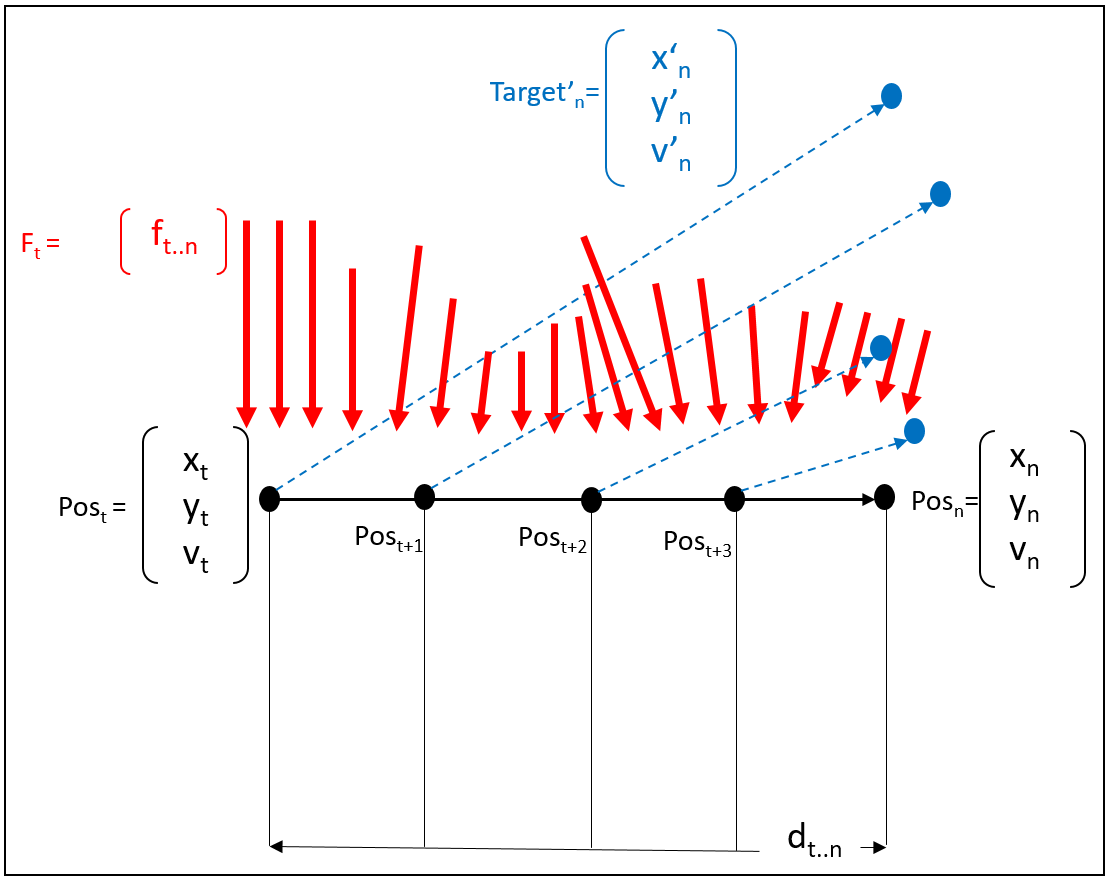}
	\caption{High\hyp level illustration of way\hyp point navigation augmentation method. Black solid line and position points denote the path we wish to maintain. Blue arrows represent the external force vector acting on the ASV, which are wind and current in our setup. Red points and arrows represent the intermediate way\hyp points provided to the Pixhawk navigator and their associated target headings.}
	\label{fig:carrot}
\end{figure}

Given an accurate model of the environment dynamics and an ability to predict temporally close external forces and their effects on the ASV, we seek to provide an augmentation to the Pixhawk way\hyp point navigation controller. By manipulating the target global pose based on the measurements and effects of external forces we are able to provide intermediate way\hyp points to the Pixhawk, coercing it to maintain the original desired trajectory; see Figure \ref{fig:carrot}. The intermediate way\hyp points account for the effects of external forces and are calculated proportional to the distance $d_t$ between the ASV and the goal way\hyp point.

\invis{IR: This sounds strange -- Other than the equations presented in earlier work, the equations to accomplish this technique are presented below.} $\textrm{Pos}_t$ is composed of the ASV's latitude, longitude, and velocity. $F_t$ is comprised of the expected effect on the ASV's speed and heading resulting from the effects models in Section \ref{subsec:effects}. $\textrm{Pos}_n$ is the goal way\hyp point and $d_t$ is the distance between the ASV and $\textrm{Pos}_n$. $\textrm{Target}'_n$ is a calculated intermediate way\hyp point to send the controller to maintain the desired trajectory. 

\invis{{\bf This is not an equation, just a list of things}
\begin{equation}
Target'_n = F\langle Pos_t, Pos_n, F_t, d_t\rangle
\end{equation}}

\begin{figure}[h]
		\centering      
		\includegraphics[width=0.99\textwidth]{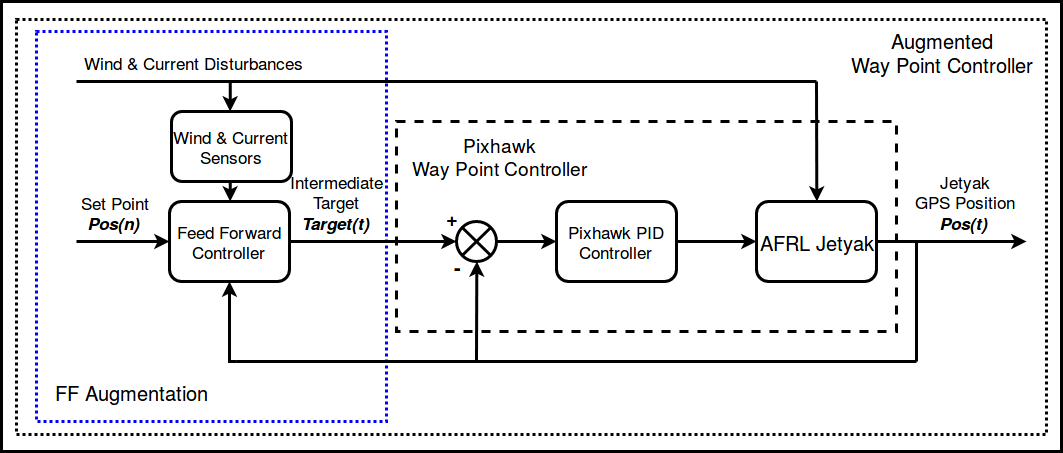}
		\caption{The way\hyp point navigation PID controller used in the Pixhawk PX4 augmented by our intermediate way\hyp point offset generator.}
		\label{fig:ff_con}
	\end{figure}

This portion of feed\hyp forward augmented controller is illustrated in Figure \ref{fig:ff_con}. The algorithm used to calculate the intermediate target way\hyp points is presented in Algorithm \ref{alg:ff}.

The inputs to the algorithm are:
\begin{itemize}
    \item The measured current speed magnitude $\textrm{spd}_c$ and direction $\textrm{dir}_c$,
    \item The measured wind speed magnitude $\textrm{spd}_w$ and direction $\textrm{dir}_w$,
    \item The ASV position ({$\textrm{lat}_t$, $\textrm{long}_t$}),
    \item The ASV speed $\textrm{spd}_t$ and heading $h_t$,
    \item The target ASV speed $\textrm{spd}\_\textrm{target}$,
    \item The list of way\hyp points in the current mission.
\end{itemize}

\invis{The inputs of the algorithm are the water and wind sensor measurements: 
the current $spd_c$ and wind $spd_w$ speed magnitudes, with their corresponding directions $dir_c$ and $dir_w$, ASV position ({$lat_t$, $long_t$}), ASV speed $spd_t$ and $h_t$, way\hyp point list ({$lat_n$, $long_n$}), target speed $spd\_target$.}

The measurements are processed as they are received from the sensors during execution of each way\hyp point from the mission. Based on the speed and orientation of the ASV we determine the absolute values of each measurement and use that to predict with a linear regression the effect of the force on the speed and direction of the ASV (Line $7-8$) \cite{moulton2019effects}. While the target way\hyp point is not reached, an intermediate way\hyp point is calculated based on the $\textrm{effect}_x$ and $\textrm{effect}_y$ values \invis{- predicted error calculated for the current way\hyp point (Line $9$)}. The speed is also adjusted based on the predicted error (Line $10$). Finally, the ASV is sent to the newly calculated way\hyp point (Line $11$). When the new target position is processed by the Pixhawk navigation controller, it results in a smoother and more accurate path, and with this we realize our original intended trajectory. In the following section, we will present the experiments carried out to demonstrate this capability. 

\begin{algorithm}
		\caption{Feed\hyp forward Augmented Way\hyp Point Navigation Controller}
		\label{alg:ff}
		\textbf{Input:} $\textrm{spd}_c$, $\textrm{dir}_c$, $\textrm{spd}_w$, $\textrm{dir}_w$, ({$\textrm{lat}_t$, $\textrm{long}_t$}), $\textrm{spd}_t$, $h_t$, way\hyp point list ({$\textrm{lat}_n$, $\textrm{long}_n$}), $\textrm{spd}\_\textrm{target}$
		\\
			\textbf{Output:} None
			\begin{algorithmic}[1]
		    \State $\textrm{mission} \gets \textrm{wp\_list}(\textrm{lat}_n, \textrm{long}_n)$
		    \State $\textrm{count} \gets |mission|$
			
			\For{\textbf{each} $\textit{i} \in \textit{1, \ldots,} \textrm{count}$}
			\State $\textrm{go\_to\_waypoint}(\textrm{lat}_i, \textrm{long}_i, \textrm{spd}\_\textrm{target})$
			\State $\textrm{wp} \gets \textrm{lat}_i, \textrm{long}_i$
			\While{$\textrm{wp}~is~not~reached$}
			\State $\textrm{effect\_spd}, \textrm{effect\_dir} \gets \textrm{effect\_model}(\textrm{spd}_c, \textrm{dir}_c, \textrm{spd}_w, \textrm{dir}_w, \textrm{spd}\_\textrm{target}, \textrm{spd}_t, h_t) $
			\State $\textrm{effect}_x, \textrm{effect}_y \gets \textrm{convert\_to\_coordinate\_vectors}(\textrm{effect}\_\textrm{spd}, \textrm{effect}\_\textrm{dir}) $
			\State $\textrm{lat}'_i, \textrm{long}'_i \gets \textrm{calc\_intermediate\_wp}(\textrm{lat}_i, \textrm{long}_i, \textrm{effect}_x, \textrm{effect}_y) $
			\State $\textrm{spd}'_i = \textrm{effect\_spd} + \textrm{spd\_target} $
			\State $\textrm{go\_to\_waypoint}(\textrm{lat}'_i, \textrm{long}'_i, \textrm{spd}'_i)$
			\EndWhile
			\EndFor
			\end{algorithmic}
\end{algorithm}

\section{Experiments}
\label{sec:exp}
\invis{In order to ensure our experiments produce meaningful results which are indicative of the utility of our work, three components are required: first a measurable hypothesis, as the one presented in prior sections; second, a reliable platform and the means to carry out numerous trials -- in Section \ref{platform} will provide details on the platform we used for testing. Finally, for the trials,} Over ten deployments were completed in support of this initiative, collecting and testing in over \SI{190}{\km} of river and lake environments; for testing the proposed controller,  four of the deployments were in the river testing the control, while the rest established a baseline behaviour and tested the effect of wind.

\subsection{Platform}
\label{platform}
The base platform is a heavily modified Mokai Es-Kape\footnote{\url{http://www.mokai.com/mokai-es-kape/}}, termed Jetyak by UofSC's Autonomous Field Robotics Laboratory, shown in Figure \ref{fig:jetyak2}. The stock boat uses an internal combustion Subaru engine and is capable of speeds up to \SI{22.5}{\km/h} and a deployment duration of over eight hours with reduced speed. The ES-Kape's factory pulse width modulated controlled servo system makes it ideal for robotic control integration.

On top of the stock configuration, this research vessel is outfitted with a Pixhawk\footnote{\url{https://docs.px4.io/en/flight_controller/mro_pixhawk.html}} flight control system. Outfitted with remote control from Taranis, long\hyp range mission tracking through RFD 900+ radios, and on\hyp board control through a companion Raspberry Pi serving to host a Robot Operating System (ROS) \cite{quigley2009ros} node, the Jetyak maintains robust architecture for autonomous initiatives as well as redundant override capabilities to ensure safe testing and experimentation.

The sensing capability is provided through NMEA 0183 depth sonar, Sparkfun anemometer for wind, and Ray Marine ST 800 paddle wheel speed sensors for current measurements. Current and wind sensors require analog to digital drivers due to the inexpensive sensing route we have selected. This is provided through Arduino Mega and Weathershield micro\hyp controllers, respectively.

\subsection{Experimental Approach}
\label{approach}

\begin{figure}[b]
\centering      
    \includegraphics[width=0.9\textwidth]{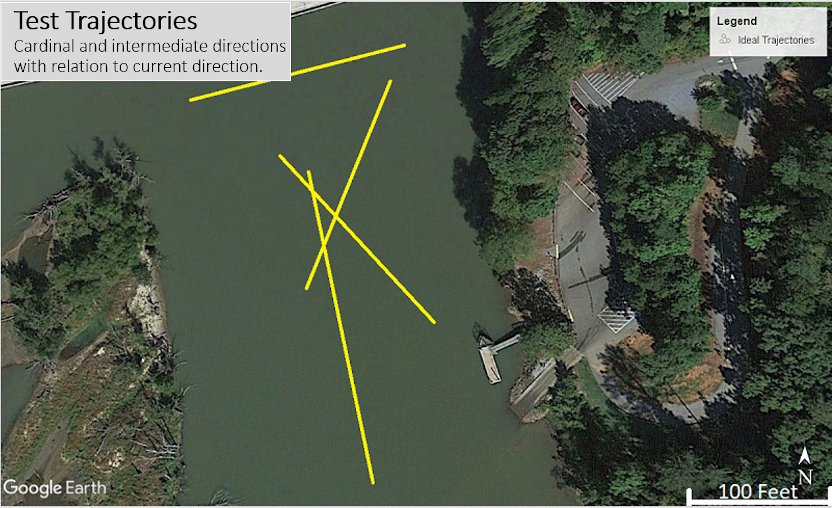}
	\caption{Test patterns run in both directions to establish a control baseline for performance evaluation in currents of less than \SI{1}{\m/s}, depending on location of the ASV in the Saluda River's cross\hyp section.}
	\label{fig:test_pattern}
\end{figure}

\invis{
\begin{wrapfigure}[14]{h}{0.5\textwidth}
\centering
\vspace{-0.3in}
\includegraphics[width=0.5\textwidth]{./figures/test_paths}
\caption{\label{fig:test_pattern}Test patterns run in both directions to establish a control baseline for performance evaluation in currents of less than \SI{1}{\m/s}, depending on location of the ASV in the Saluda River's cross\hyp section.}
\end{wrapfigure}
}
In order to provide experimental results that are easily comparable to the original
way\hyp point PID controller, we use straight line test trajectories that run in the cardinal directions parallel and perpendicular to the predominant external force.
In this case, currents are being tested, and we illustrated in Figure \ref{fig:problem} that the Jetyak's poorest path following performance occurs when travelling in the same direction as the current. This led us to select the four cardinal and four intermediate direction orientations to the current as our test baseline, shown in Figure \ref{fig:test_pattern}. Straight line segments were produced to replicate the most common patterns from route planning experiments. The generated segments were initially used as input to the standard Pixhawk way\hyp point controller. Then, the same segments were used as input to the augmented controller with the intermediate way\hyp points enabled.

Given this controlled experimental setup, the results in Section \ref{sec:results} illustrate the success of this approach as well as directions for future work.

\section{Results}
\label{sec:results}
In this section we will compare the performance of the standard Pixhawk GPS way\hyp point navigation controller with and without the proposed feed\hyp forward augmentation. Since the standard controller performs well in upstream maneuvers, the focus will be on the performance difference in the downstream cases. Due to weather constraints during the field trials, the results presented in this paper were obtained from data collected in a river with an average measured speed of \SI{0.677}{\m/s} during the trial for straight trajectories. \invis{Improving turning accuracy in external forces will be discussed in the future work portion of this paper.}
\begin{figure*}[ht]
		\centering
		\leavevmode
		\begin{tabular}{cc}
		   \includegraphics[width=0.49\textwidth]{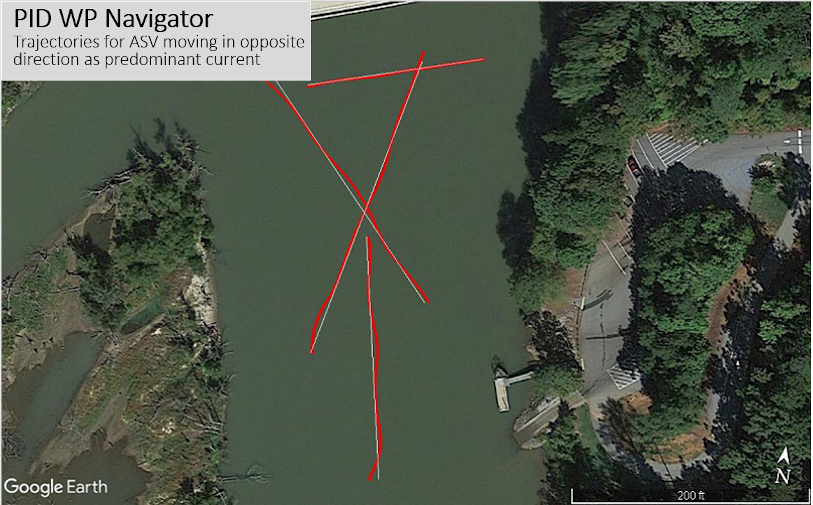}\label{fig:wp_nav_against}&
		    \includegraphics[width=0.49\textwidth]{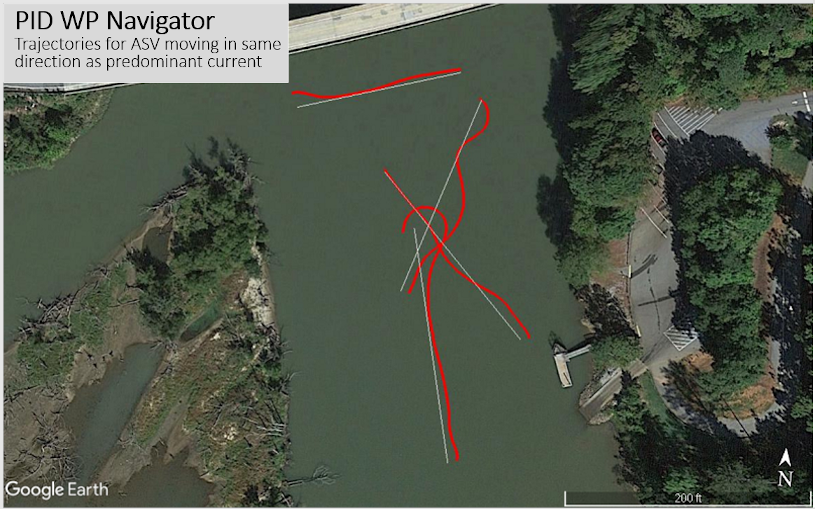}\label{fig:wp_nav_with}
		\end{tabular}
			\caption{Pixhawk PID controlled way\hyp point navigator tracking in slow currents with the ASV travelling mainly (a) against the predominant direction of the current; (b) with the predominant direction of the current  -- white line: target trajectory, red line: actual executed trajectory.
			\label{fig:wp_nav_results}}
	\end{figure*}

\subsection{Way\hyp point Navigation}
\label{subsec:wp_nav_results}
As illustrated in Figure \ref{fig:wp_nav_results}, the built\hyp in Pixhawk controller is generally able to reach the required way\hyp points. However, the PID coefficients are tuned to operate in a specific environment. When changing environments, the PID coefficients should be tuned again. This task becomes insurmountable when operating in environments with ever\hyp changing dynamic forces at play. As shown in Figure \ref{fig:wp_nav_results} left, negotiating currents in upstream to perpendicular directions is relatively stable. This is due to the fact that the speed of the ASV relative to the ground is slightly reduced, allowing enough time for the PID controller to compensate for the error. However, in Figure \ref{fig:wp_nav_results} right, we see the opposite effect when the speed of the ASV relative to the ground is increased, thereby accumulating too much error in the PID controller to overcome the external forces. This typically results in an overshoot scenario where the ASV begins harmonically oscillating back and forth over the desired trajectory. It should also be noted, that as the speed of the current increases, this behavior starts to present itself in trajectories perpendicular to the current. Adjusting the integral gain in the PID controller can help solve this problem, but it will also produce undesirable oscillatory behavior in upstream trajectories.

\begin{figure*}[ht]
		\centering
		\leavevmode
		\begin{tabular}{cc}
		   \includegraphics[width=0.49\textwidth]{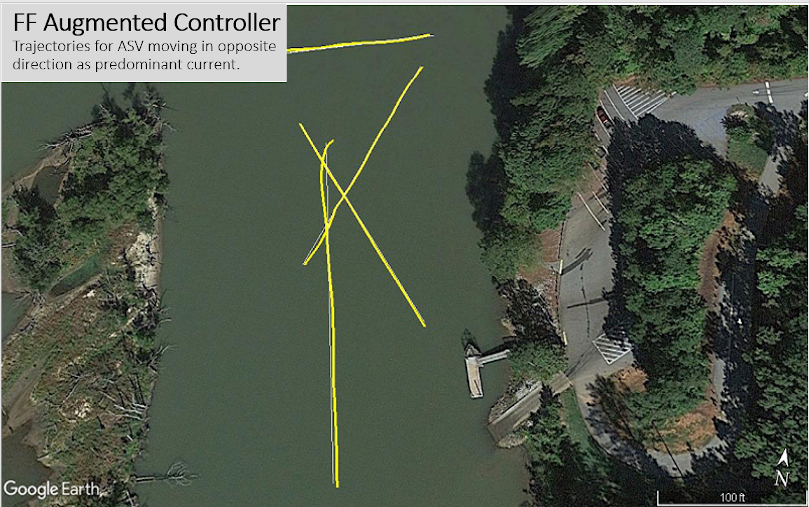}\label{fig:augmented_against}&
			\includegraphics[width=0.49\textwidth]{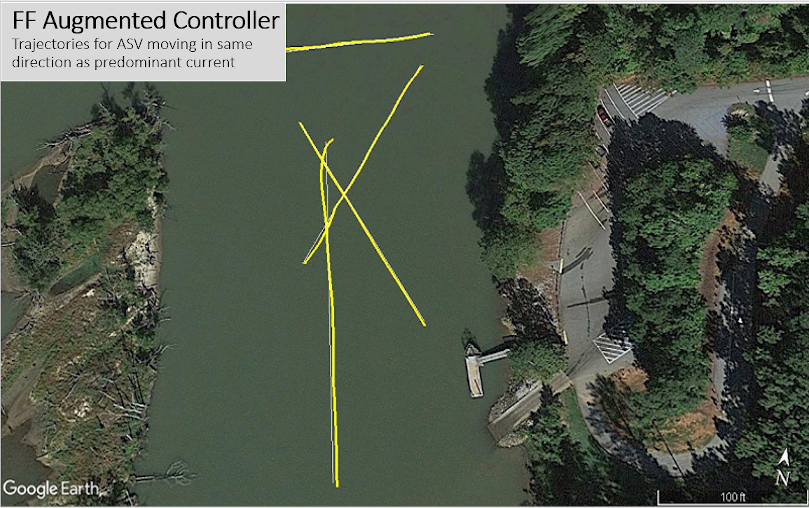}\label{fig:augmented_with}
		\end{tabular}
			\caption{Augmented Pixhawk way\hyp point navigator tracking in slow currents with the ASV travelling mainly (a) against the predominant direction of the current; (b) with the predominant direction of the current -- white line: target trajectory, yellow line: actual executed trajectory.
			\label{fig:augmented_results}}
	\end{figure*}

\subsection{Proactive Effects Augmented Way\hyp point Navigation Controller}
\label{subsec:dynamic_results}

As illustrated in Figure \ref{fig:augmented_results}, by augmenting the built-in PID controller in the Pixhawk, we were able to follow much more precisely the desired path to each way\hyp point than the non\hyp augmented controller. These results serve as proof of concept for Algorithm \ref{alg:ff}. 
As shown in Figure \ref{fig:augmented_results}, path following in currents in all orientations to the ASV is qualitatively improved.  

\begin{table}[]
\caption{Comparing the performance of the standard Pixhawk way\hyp point PID controller with our intermediate way\hyp point augmented control. Perpendicular results represent average error over two traversals in opposing directions. \invis{Max error represents the largest distance between the straight\hyp line trajectory and the actual path of the ASV. Percentage path error greater than one meter quantifies the portion of the path where the ASV was more than one meter from the ideal trajectory.}}
\label{table:compare}
\begin{tabular}{cc|c|c|c|c|c|c|c|}
\cline{3-9}
\multicolumn{2}{c}{}                                                                                                                                                                                   & \multicolumn{7}{|c|}{ASV Trajectory Relative to Current}                                                                                \\ \cline{3-9} 
\multicolumn{2}{c|}{}                                                                                                                                                                                   & \rotatebox{90}{Perpendicular} & \rotatebox{90}{Parallel With} & \rotatebox{90}{Parallel Against} & \rotatebox{90}{L-R Diagonal With} & \rotatebox{90}{L-R Diagonal Against} & \rotatebox{90}{R-L Diagonal With} & \rotatebox{90}{R-L Diagonal Against} \\ \hline
\multicolumn{1}{|c|}{\multirow{2}{*}{\begin{tabular}[c]{@{}c@{}}WP Navigator\\ with PID Control\end{tabular}}}              & \begin{tabular}[c]{@{}c@{}}Max\\ Error\end{tabular}                    & 4.50 m        & 9.32 m        & 3.86 m           & 3.46 m            & 1.63 m               & 7.85 m            & 2.57 m               \\ \cline{2-9} 
\multicolumn{1}{|c|}{}                                                                                                      & \begin{tabular}[c]{@{}c@{}}\% Path \\ Error \textgreater 1 m\end{tabular} & 44.2\%        & 76.8\%        & 18.3\%           & 48.6\%            & 12.7\%               & 83.5\%            & 40.1\%               \\ \hline
\multicolumn{1}{|c|}{\multirow{2}{*}{\begin{tabular}[c]{@{}c@{}}Augmented WP \\ Navigator with\\ PID Control\end{tabular}}} & \begin{tabular}[c]{@{}c@{}}Max\\ Error\end{tabular}                    & 1.58 m        & 1.48 m        & 1.08 m           & 0.75 m            & 0.74 m                & 1.07 m            & 0.68 m               \\ \cline{2-9} 
\multicolumn{1}{|c|}{}                                                                                                      & \begin{tabular}[c]{@{}c@{}}\% Path \\ Error \textgreater 1m\end{tabular} & 9.3\%         & 11.9\%        & 7.9\%            & 0\%               & 0\%                  & 6.3\%             & 0\%                  \\ \hline
\end{tabular}
\end{table}

The results in Table \ref{table:compare} show a quantitative comparison of the performance of our augmented proactive controller with the baseline way\hyp point navigator. In particular, a marked improvement can be observed in both maximum error and percentage of the path that is more than a meter far from the target trajectory. Max error represents the largest distance between the straight\hyp line trajectory and the actual path of the ASV. Percentage path error greater than one meter quantifies the portion of the path where the ASV was more than one meter from the ideal trajectory. Confirming intuition, the ability of the augmented control algorithm to change the forward thrust of the ASV provides the largest numerical improvement when moving with the predominant direction of the current. \invis{jason- not what i meant to say. It should be noted, however, that the addition of intermediate way\hyp points improves path following for all orientations between the current and the ASV direction of travel. }

\section{Conclusions}
\label{sec:conclusion}

The path\hyp following precision achieved by this work can have profound impacts for the research, emergency services, and exploration communities. \invis{ For instance, the scenario immediately following the Fukushima nuclear disaster could have been more quickly visited, negotiated, and assessed possibly saving lives or providing strategies for containment of the site.} The ability to provide bathymetric surveying and mapping capabilities to remote areas with highly dynamic currents will enable researchers to expand the boundary between known and unknown environments.

To improve the robustness of the control augmentation presented, future work should include two areas. First,\invis{full integration of all modules of the augmenting feed\hyp forward algorithm should be self-contained and transparent to operators and the Pixhawk mission planner.} the addition of providing the same precision path following for a Dubin's vehicle, such as the Jetyak will require additional methods to handle deliberate turns in planned missions. Second, another desirable expansion of this work will include changing from intermediate way\hyp point augmentation to a lower level control of the linear and angular velocities ($v, \omega$). Such an approach may produce more concise countermeasures to further reduce the path tracking error.  

%
%

\begin{acknowledgement}
The authors would like to thank the generous support of the National Science Foundation grants (NSF 1513203, 1637876).
\end{acknowledgement}
%
%
%

\bibliographystyle{template/spmpsci}
\bibliography{author} 
\end{document}